\pdfoutput=1

\documentclass[11pt]{article}

\usepackage[final]{acl}

\usepackage{times}
\usepackage{latexsym}

\usepackage[T1]{fontenc}

\usepackage[utf8]{inputenc}

\usepackage{microtype}

\usepackage{inconsolata}

\usepackage{graphicx} 

\usepackage[textsize=small]{todonotes}
\usepackage{amsmath}
\usepackage{amssymb}

\usepackage{algorithm}
\usepackage{float}
\usepackage{cuted}
\usepackage{multirow}
\usepackage{longtable}
\usepackage{listings}
\usepackage[noend]{algpseudocode}
\usepackage[T1]{fontenc}
\usepackage[english]{babel}
\usepackage[utf8]{inputenc}
\usepackage{tabularx}
\usepackage{siunitx}

\title{Developing PUGG for Polish: A Modern Approach to KBQA, MRC, and IR Dataset Construction}

\author{Albert Sawczyn \And  Katsiaryna Viarenich \And Konrad Wojtasik \AND Aleksandra Domogała \And Marcin Oleksy \And Maciej Piasecki \AND Tomasz Kajdanowicz \AND
        \vspace{-1em} \\ Wrocław University of Science and Technology \\ albert.sawczyn@pwr.edu.pl}

\begin{document}
\maketitle
\begin{abstract}
Advancements in AI and natural language processing have revolutionized machine-human language interactions, with question answering (QA) systems playing a pivotal role. The knowledge base question answering (KBQA) task, utilizing structured knowledge graphs (KG), allows for handling extensive knowledge-intensive questions. However, a significant gap exists in KBQA datasets, especially for low-resource languages. Many existing construction pipelines for these datasets are outdated and inefficient in human labor, and modern assisting tools like Large Language Models (LLM) are not utilized to reduce the workload. To address this, we have designed and implemented a modern, semi-automated approach for creating datasets, encompassing tasks such as KBQA, Machine Reading Comprehension (MRC), and Information Retrieval (IR), tailored explicitly for low-resource environments. We executed this pipeline and introduced the PUGG dataset, the first Polish KBQA dataset, and novel datasets for MRC and IR. Additionally, we provide a comprehensive implementation, insightful findings, detailed statistics, and evaluation of baseline models.
\end{abstract}

\section{Introduction}
\label{sec:introduction}

Question answering (QA) systems serve as a sophisticated interface between humans and computers. To further enhance their utility, we need QA systems capable of answering questions based on extensive knowledge \citep{petroniKILTBenchmarkKnowledge2021}.
The knowledge base question answering (KBQA) task addresses this need by using structured knowledge graphs (KG) to provide accurate and relevant answers \cite{Lan2021survey}. KBQA leverages these graphs, which are rich with interconnected entities and relationships, to decode complex queries and deliver precise answers. Importantly, systems that reason over KGs are more resistant to the phenomenon of hallucinations, common in large language models (LLM) \cite{baek-etal-2023-knowledge-augmented}. Additionally, the inherent flexibility of KGs facilitates easy modification and updating, ensuring the use of only the most current and accurate facts.

However, a significant gap exists in KBQA datasets. Many are schematic and not natural in their language, or they rely on discontinued knowledge graphs \citep{Lan2021survey, Steinmetz2021, Jiang2022kg}. By \textit{natural} we refer to naturally occurring questions \cite{kwiatkowski-etal-2019-natural}.  While a broader range of KBQA datasets is available for English, most low-resource languages, including Polish, lack such resources \cite{rubq1.0}. This scarcity is part of a broader issue prevalent in the field of NLP concerning low-resource languages \cite{lepiszcze}. Recognizing this gap, we set out to create a KBQA dataset for Polish. We faced several challenges during extensive studies of existing works to find the most efficient methods for dataset creation. Many datasets were built on simpler predecessors \cite{rubq1.0, kaffee_et_al:TGDK.1.1.10}, and also many construction pipelines are inefficient regarding human labor, as they do not utilize modern tools that could reduce human work, such as assisting Large Language Models (LLM). LLMs have opened new opportunities for assisting human annotators, especially in low-resource languages where the range of pre-trained models is limited \citep{doi:10.1073/pnas.2305016120, kuzman2023chatgpt}. 

Consequently, we decided to design, implement, and execute a modern approach to creating KBQA datasets tailored explicitly for the low-resource environment. We selected Wikidata as KG due to its extensive, multilingual coverage and dynamic, open, and free nature \cite{10.1145/2629489}. Notably, we did not use any translation, ensuring that the output data sounded natural. Moreover, an advantageous byproduct of this pipeline was the concurrent development of machine reading comprehension (MRC) and information retrieval (IR) datasets, requiring no extra human annotation. MRC is essential for AI to understand and analyze texts like a human reader \cite{rajpurkar-etal-2016-squad, kwiatkowski-etal-2019-natural}, while IR is crucial for efficiently extracting relevant information from vast databases \cite{nguyen2017ms, thakur2021beir}.

We summarize our contributions as follows:

\begin{itemize}
    \item We introduce the PUGG \footnote{The name "PUGG" refers to "Pirate Pugg", a fictional character from "The Sixth Sally" of "The Cyberiad" by Stanisław Lem. Pirate Pugg is depicted as being obsessed with gathering information.} dataset, which encompasses three tasks --- KBQA, MRC, and IR\footnote{\url{https://huggingface.co/datasets/clarin-pl/PUGG}}. This dataset features natural factoid questions in Polish and stands out as the first Polish KBQA resource \footnote{The dataset license: CC BY-SA 4.0}. To address a range of complexities, we have enriched the dataset by complementing natural questions with simpler, template-based questions.
    \item We propose a semi-automated dataset construction pipeline designed for low-resource environments. The pipeline results in the creation of KBQA, MRC, and IR datasets while drastically reducing the labor of human annotators. Accompanying this pipeline is a comprehensive implementation \footnote{\url{https://github.com/CLARIN-PL/PUGG}}. Moreover we share insightful findings and detailed statistics obtained from the PUGG dataset construction using the pipeline. These provide valuable resources for future developers of datasets. Additionally, we developed few utility custom methods, e.g. for entity linking, that are useful in diverse contexts.
    \item We provide an evaluation of baseline models, thereby establishing benchmarks for future research using the PUGG dataset.
\end{itemize}

\section{Related Work}
\label{sec:related-work}

\begin{figure*}[!htb]
    \centering
    \includegraphics[width=0.95\textwidth]{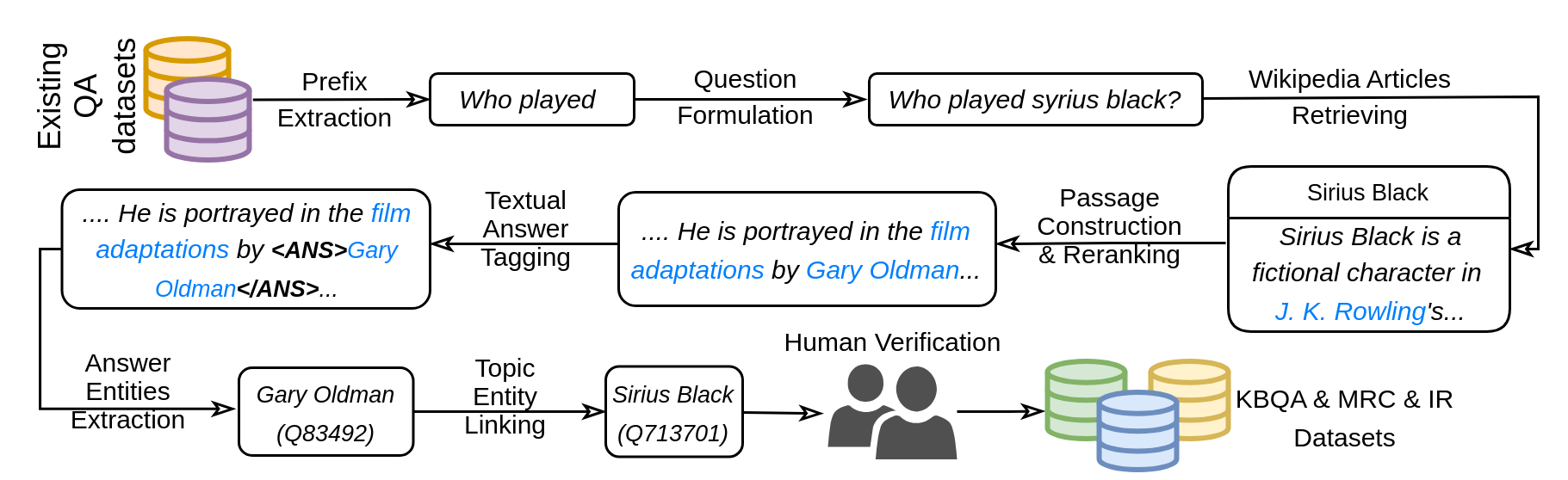}
    \caption{\label{fig:pipeline}
    Overview of the proposed construction pipeline for natural questions. The figure shows the processing of a single example. Rounded rectangles represent acquired data, with blue text indicating a hyperlink to another Wikipedia article. Arrow descriptions indicate automated procedures. The symbol of people denotes a step involving human verification depicted in  Section \ref{sec:pipeline-human-ver}: Human Verification and in Figure \ref{sec:pipeline-human-ver}. The example data is in English for non-Polish readers, but the pipeline was originally executed on Polish data for PUGG creation.  
    }
\end{figure*}

\paragraph{KBQA}

Existing KBQA datasets have been comprehensively studied and compared in works done by \citet{rubq1.0} and \citet{Jiang2022kg}. A significant finding is the lack of a Polish KBQA dataset. Most KBQA datasets are primarily in English, with exceptions like the Chinese NLPCC-KBQA \cite{10.1007/978-3-319-73618-1_86}, Russian RuBQ \cite{rubq1.0}, the multilingual QALD \cite{9736271} and MCWQ \cite{10.1162/tacl_a_00499} (both not including Polish). The closest dataset resembling a KBQA task in Polish is the multilingual MKQA \cite{longpre-etal-2021-mkqa}, where approximately 42\% of its 10,000 questions are answerable by Wikidata entities. However, MKQA cannot be classified as a proper KBQA dataset due to the lack of annotated topic entities.

The study by \citet{rubq1.0} outlines the various question generation techniques used in existing KBQA datasets. For generating natural questions in our research, we adopted a question generation technique based on query suggestion, initially introduced by \citet{berant-etal-2013-semantic}. This technique is effective for acquiring natural factoid questions likely to be posed to a QA system, similar to the approaches used in datasets like NQ \cite{kwiatkowski-etal-2019-natural} and WikiQA \cite{yang-etal-2015-wikiqa}, which were built from questions asked to search engines. For template-based questions, our approach involved creating questions from predefined reasoning templates, a standard method in many KBQA datasets \cite{bordes2015largescale, su-etal-2016-generating, 10.1007/978-3-030-30796-7_5}. Several KBQA datasets used crowdsourced paraphrasing for question diversification \cite{talmor-berant-2018-web, su-etal-2016-generating, 10.1007/978-3-030-30796-7_5}. In contrast, our approach only automates this process by incorporating humans during final verification.

\paragraph{IR}
Many valuable resources for Information Retrieval in the Polish language were recently created. The BEIR-PL \cite{wojtasik2023beirpl} benchmark was proposed as an automatic machine translation of the BEIR \cite{thakur2021beir} benchmark. This popular zero-shot retrieval benchmark was originally only for the English language. The MQUPQA \cite{rybak-2023-maupqa} dataset is a composition of multiple already existing Polish and multilingual datasets, like CzyWiesz \cite{czywiesz}, MKQA \cite{longpre-etal-2021-mkqa}. Additionally, the MQUPQA dataset incorporates other automatic methods for question and answer generation, such as utilizing the generative capabilities of the GPT-3 model \cite{gpt3} or employing templates inspired by the structure of Wikipedia. The PolEval \cite{FedCSIS20235627} competition featured a passage retrieval task. It comprised three datasets from various domains: Wikipedia-based, e-commerce shop FAQ, and legal questions.
Currently, a Polish Information Retrieval Benchmark (PIRB) \cite{dadas-etal-2024-pirb-comprehensive} provides a platform to evaluate models across various datasets. The models referred to in this benchmark represent the current state-of-the-art in Polish IR.

\paragraph{MRC}
 QA datasets often have a close relationship with IR datasets. The CzyWiesz dataset is based on the \textit {Did you know?} section of Wikipedia, with provided answers and also relevant articles. Another example is the PolQA \cite{rybak2022improving} dataset, which is comprised of general questions from quiz shows annotated with relevant passages from Wikipedia. The PoQuAD \cite{poquad} dataset is structured around questions manually annotated to correspond with the best articles on Wikipedia, mirroring the methodology of the SQuAD \cite{rajpurkar-etal-2016-squad} dataset. Contrastively, our dataset consists of naturally occurring questions, which are afterward annotated to relevant articles.

\section{Definitions}
\label{sec:task-descriptions}

A common element in the tasks of KBQA, MRC, and IR is the textual question $q$. We denote the set of questions as $\mathcal{Q}$.  Despite \textit{query} being common in the field of IR, we use \textit{question} and \textit{query} interchangeably, as our dataset's queries take the form of questions.

\paragraph{KBQA} 
We denote KG as a multi-relational heterogeneous graph $\mathcal{G}=(\mathcal{E}, \mathcal{R}, \mathcal{T})$, composed of three elements: a set of entities $\mathcal{E}$, a set of relation predicates $\mathcal{R}$, and a set of triples (facts) $\mathcal{T}$. Each triplet $(h, r, t) \in \mathcal{T}$ indicates a relation predicate $r$ between two entities, a head entity $h$ and a tail entity $t$, where $h, t \in \mathcal{E}$ and $r \in \mathcal{R}$ \cite{hamiltonRepresentationLearningGraphs2017}. In the KBQA task, a textual question $q$ and associated topic entities $\mathcal{E}_{q} \subset \mathcal{E}$  are given. The objective is to retrieve answer entities $\mathcal{A}_q\subset\mathcal{E}$ that satisfy the question based on facts in the $\mathcal{G}$. Therefore, we denote KBQA dataset as $\mathcal{D}_{KBQA} = \{(q, \mathcal{E}_{q}, \mathcal{A}_q) \}$.

\paragraph{MRC} 
MRC aims to answer a textual question \( q \) based on a given text passage \( p_q \). We denote MRC dataset as \( \mathcal{D}_{MRC} = \{(q, p_q, a_q)\)\}, where \( a_q \) is the answer extracted from \( p_q \).

\paragraph{IR}
The IR task focuses on finding a passage $p$ from a large corpus relevant to a query \( q \). The corpus \( \mathcal{C} \) is defined as a set of passages, i.e., \( \mathcal{C} = \{p_1, p_2, ..., p_n\} \). The IR dataset is denoted as \( \mathcal{D}_{IR} = \{(q, p_q)\)\}, where \( p_q \in \mathcal{C} \) denotes a passage that is relevant to the query \( q \).

\section{Construction Pipeline}
\label{sec:pipeline}

This section introduces the construction pipeline for the PUGG dataset, specifically designed to create a dataset with natural and factoid questions in a semi-automated manner. This approach significantly reduces the workload of human annotators. We outline the pipeline's fundamental design, presented in Figure \ref{fig:pipeline}, emphasizing its adaptability to various environments. While this part focuses on the general framework, specific implementation details, such as the models and algorithms used, will be discussed in Section \ref{sec:pipeline-execution}.

\paragraph{Question Formulation} The initial step of our pipeline involves acquiring a variety of natural factoid questions. We initiated our process using existing datasets to minimize the need for manual annotation. From previously existing QA datasets, we extract question prefixes ranging from basic (\textit{'who...'}, \textit{'when...'}) to more specific (\textit{'which Canadian athlete...'}, \textit{'which theater co-created...'}). Then, the gathered prefixes are completed to formulate a set of questions. We can employ various methods, including rule-based approaches and language models \cite{Das2021}, and for natural questions, we can also integrate external services.

At this stage, we have a collection of question candidates $q'$, as some of which may be incorrect. These inaccuracies are not a concern at this point, as they will be filtered out during the human verification process, detailed in Section \ref{sec:pipeline-human-ver}.

\paragraph{Passage Construction} The next phase involves text passages retrieval to answer the formulated questions. We use a data source with referenced graph entities, which in our case is Wikipedia. To find relevant articles for each question, various retrieval techniques can be employed, such as dense retrieval \cite{reimers-gurevych-2019-sentence} with additional reranking. Once relevant articles are identified, they are segmented into smaller passages and reranked to prioritize passages most likely to contain an answer. 

All passages constructed in this phase are added to the passage corpus $\mathcal{C}$ needed for the IR task.

\paragraph{Textual Answers, Answer Entities} 
We select the most accurate passage as candidate passage $p_{q}'$, and we apply a QA model, such as LLM or pre-trained extractive model, to tag a span of passage denoting a candidate textual answer $a_{q}'$. Such textual answers contain hyperlinks to other articles associated with specific Wikidata entities. We extract these entities and build a set of candidate answer entities $\mathcal{A}_{q}'$. 

\paragraph{Topic Entities} The subsequent step in our pipeline is performing an entity linking process to identify and link the KG entities mentioned in the questions. We refer to them as candidate topic entities $\mathcal{E}_{q}'$.

\begin{figure}[!htb]
    \centering
    \includegraphics[width=0.60\linewidth]{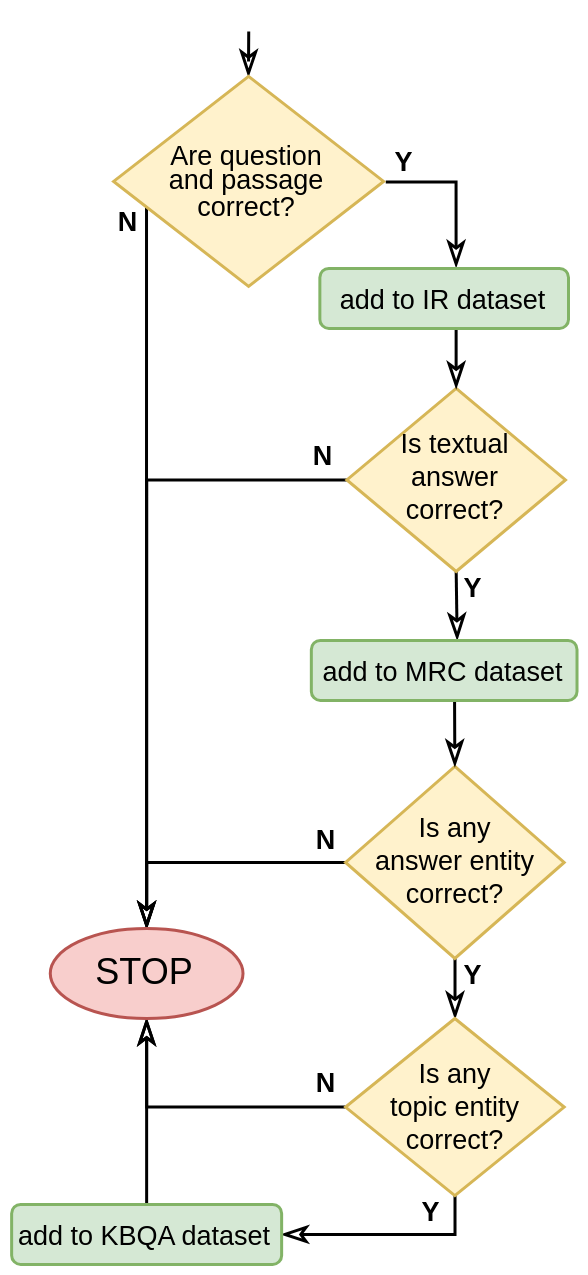}
    \caption{\label{fig:human-ver}
    The human verification procedure for all acquired candidates.
    }
\end{figure}

\paragraph{Human Verification} \label{sec:pipeline-human-ver}
To this point, we have acquired all necessary data to construct the KBQA, MRC, and IR datasets: questions $q'$ accompanied by a passage $p_{q}'$, textual answer $a_{q}'$, answer entities $\mathcal{A}_{q}'$, and topic entities $\mathcal{E}_{q}'$. All these elements are obtained through fully automated processes. While automation significantly reduces the need for human labor, it is not entirely error-proof. To ensure the high quality of our dataset, we implement a human verification process. The detailed procedure of this human verification is depicted in Figure \ref{fig:human-ver}. During this process, candidate elements  $q'$, $p_{q}'$, $a_{q}'$, $\mathcal{A}_{q}'$, and $\mathcal{E}_{q}'$ undergo verification. This leads to the final elements $q$, $p_q$, $a_q$, $\mathcal{A}_q$, and $\mathcal{E}_{q}$, respectively. 
The final sets $\mathcal{A}q \subseteq \mathcal{A}_{q}'$ and $\mathcal{E}_{q} \subseteq \mathcal{E}_{q}'$ indicate that the validated entities are subsets of their initial candidate sets. Note that the verification procedure (Figure \ref{fig:human-ver}) consists of multiple conditions, which may result in the datasets varying in size. This is reflected in the relationship $|\mathcal{D}_{IR}| \geq |\mathcal{D}_{MRC}| \geq |\mathcal{D}_{KBQA}| $.

\paragraph{Template-based KBQA}
While the proposed pipeline generates natural questions, we also created template-based questions to enrich our dataset. We wanted to provide a broader training and evaluation platform by offering a more schematic and straightforward set of questions, ensuring a reasoning path between topic and answer entities. The template-based questions are also beneficial for semantic parsing-based KBQA methods \cite{Lan2021survey}. 

\begin{figure*}[!htb]
    \centering
    \includegraphics[width=\textwidth]{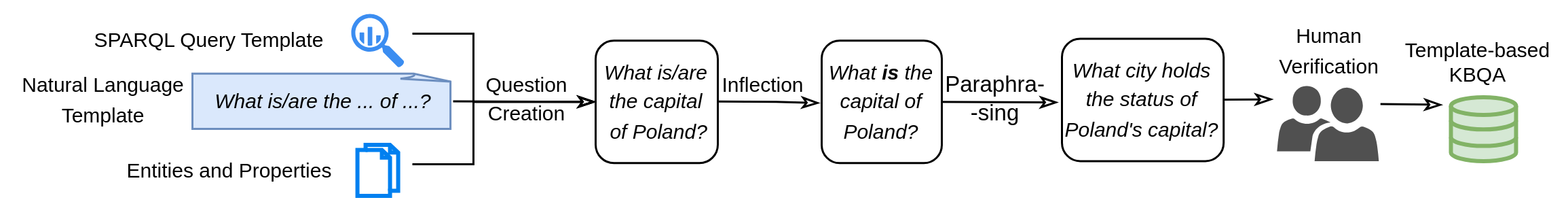}
    \caption{\label{fig:pipeline-template}
    Overview of the proposed construction pipeline for template-based questions. The figure shows the processing of a single example. The symbol of people denotes a step involving human verification to ensure all questions are meaningful. The example data is in English for non-Polish readers, but the pipeline was originally executed on Polish data for PUGG creation.
    }
\end{figure*}

Figure \ref{fig:pipeline-template} depicts the procedure of creating template-based questions. We create SPARQL templates paired with corresponding natural language questions, representing specific reasoning paths in the KG. We specify potential entities and relations to be used within these templates. We insert these entities and relations into the natural language template to construct questions. Then, we run the corresponding SPARQL queries to retrieve answer entities.

At this stage, the formulated questions might sound unnatural, especially in inflected languages like Polish. We use two strategies to address this: word inflection and question paraphrasing. We can automate the inflection process using NLP tools like spaCy \cite{honnibal2020spacy} or LLMs. We also use LLMs to paraphrase the questions for greater diversity and complexity. Given the automation of these processes, we ensure the meaningfulness of all questions through human verification.

\section{Pipeline Execution}
\label{sec:pipeline-execution}

This section delves into the specific implementation of the construction pipeline for the PUGG dataset, as previously outlined in a general framework in Section \ref{sec:pipeline}.  Our implementation was adapted for Polish NLP resources, which face challenges like limited task-specific pre-trained models and lower performance than English. 

\paragraph{Question Formulation}  In implementing our question acquisition step, we utilized two Polish datasets, CzyWiesz \cite{czywiesz} and PoQuAD \cite{poquad}. Question prefixes were extracted either by taking the first $\{1,2,3\}$ tokens from each question or by extracting text up to the first occurrence of a named entity. We employed three NER models: $pl\_core\_news\_sm$, $pl\_core\_news\_lg$ from Spacy \cite{honnibal2020spacy}, and WikiNEuRal \cite{tedeschi-etal-2021-wikineural-combined}. Each of these models provided a unique perspective in identifying named entities, thereby contributing to the variety of the prefixes. To formulate natural questions from these prefixes, we followed previous studies \cite{berant-etal-2013-semantic, 10.1007/978-3-030-77385-4_32} and used the Google Suggest API.

\paragraph{Passage Construction} We followed established methodologies from prior research \cite{kwiatkowski-etal-2019-natural} and employed the Google Search Engine \footnote{\url{https://developers.google.com/custom-search/v1/overview}} to retrieve Wikipedia articles relevant to each question. We processed the top 10 search results using the API, focusing on Wikipedia entries. Questions without a Wikipedia article in the top 10 results were discarded. The text and inter-article references of these Wikipedia articles were then obtained using the Wikipedia API\footnote{\url{https://pl.wikipedia.org/w/api.php}}. The retrieved articles were segmented into shorter passages using a sliding window approach, with a window length of $120$ words and a step size of $60$ words. We ranked these passages for each question according to their relevance. This was achieved by leveraging the PyGaggle \cite{pygaggle} library with the multilingual reranker model \mbox{\textit{unicamp-dl/mt5-3B-mmarco-en-pt}} \cite{bonifacio2021mmarco}.

\paragraph{Textual Answers, Answer Entities} For textual answer tagging, we employed \textit{GPT-3.5-turbo} \footnote{\url{https://platform.openai.com/docs/models/overview}} \cite{gpt3} with an originally designed prompt, detailed in Appendix \ref{sec:appendix-tagging}. Due to the model's generative nature and tendency to alter or paraphrase the original text, we developed a custom method to extract tagged segments accurately. This method is described in Appendix \ref{sec:appendix-tagging}. As previously described, candidate answer entities were directly referenced in the text, allowing for straightforward extraction.

\paragraph{Topic Entities} Implementing the entity linking step presented several challenges, primarily due to the lack of robust tools or models for entity linking in the Polish language. Our testing of multilingual models like mGENRE \cite{decao2020multilingual} and adapted for Polish BLINK \cite{wu2019zero} yielded unsatisfactory results, particularly for short contexts such as individual questions. Additionally, given the planned human verification stage, a method with high recall was desired. To address these challenges, we developed a heuristic method tailored to our requirements and the available resources. It leverages the Wikipedia search engine to identify potential entities for single words, combined words and identified named entities. Additionally, it collects entities referenced in the retrieved pages and utilizes title similarity measures to ensure the relevance of identified entities to the question. More details on the method can be found in Appendix \ref{sec:appendix-entity-linking}.

\paragraph{Human Verification} \label{sec:pipeline-execution-human-ver-text-ans}

The general procedure for human verification is illustrated in Figure \ref{fig:human-ver}. We implemented this by dividing it into two distinct stages. The first stage focused on identifying two aspects: questions with correctly assigned passages and questions where the textual answers within these passages were accurately tagged. The second stage of human verification had two parts: annotators marked the correct answer entities and then identified the correct topic entities. All annotators were employed in Poland and fluent in Polish. They were familiar with the Polish culture and social context. Appendix \ref{sec:appendix-human-verification} presents more details about annotation procedures and guidelines.

\paragraph{Template-based KBQA}

The developed templates are detailed in Appendix \ref{sec:appendix-templates}. It is important to note that while our template-based KBQA dataset contains fewer templates compared to other datasets, ours are more general. This is achieved by injecting not only entities but also relations into the templates, enhancing their diversity. We used entities from Wikipedia's Vital Articles Level 4 \footnote{\url{https://en.wikipedia.org/wiki/Wikipedia:Vital_articles}} and $173$ manually selected relations. Any entities lacking a Polish label were excluded. Given the vast number of possible inputs (entities and relations) for the templates and that most of them will not yield answers, random input selection was not feasible. Therefore, we divided the process into two steps, each involving the execution of a SPARQL query. First, we gathered potential sets of inputs, and then, we selected some of these sets to retrieve answers. We also utilized the specified inputs to create questions using natural language templates.

Then, we inflected and paraphrased the constructed questions using the \textit{GPT-3.5-turbo} model \cite{gpt3}. Following this, we filtered out examples without high similarity to their original form based on the longest common sequence analysis. One annotator verified the questions. Similarly to the natural questions, the annotator was employed in Poland, fluent in Polish, and familiar with Polish culture and social context. The statistics of the verification can be found in Appendix \ref{sec:appendix-templates}.

\paragraph{Outcome} The execution of our pipeline resulted in the creation of the PUGG dataset, featuring three tasks: KBQA (natural and template-based), MRC, and IR. Statistics for each dataset are presented in Table \ref{tab:statistics}. The detailed statistics of the pipeline steps, unique entities and relations in the dataset, and the distribution of examples across different template types are available in \ref{sec:appendix-detailed-stats}. Due to the utilized sliding window approach in passage construction, all passages from corpus $\mathcal{C}$  that overlapped with any of $p_c$ were removed. As Wikidata is a vast KG and using it for research can be inconvenient, we provide sampled versions of the KG: Wikidata1H and Wikidata2H. These are subgraphs created by traversing $1$ or $2$ relations from each answer and topic entity, representing two different levels of data complexity.

\begin{table}[!htb]
\centering
\small
\begin{tabularx}{\linewidth}{XlS[table-format=6.0]}
\hline \hline

            \textbf{Dataset} & \textbf{} &  \textbf{Size} \\
\hline
\hline
       \multirow{2}{*}{\textbf{KBQA} (natural)} &  \textit{train} &  2776 \\
        &   \textit{test} &   695 \\
\hline
      total  &    &   3471 \\
\hline
\multirow{2}{*}{\textbf{KBQA} (template-based)} &  \textit{train} &  1697 \\
                                       &   \textit{test} &   425 \\
\hline
      total  &    &   2122 \\
\hline
\multirow{2}{*}{\textbf{KBQA} (all)} &  \textit{train} &  4473 \\
                                       &   \textit{test} &   1120 \\
\hline
      total  &    &   5593 \\
\hline
\hline
       \multirow{2}{*}{\textbf{MRC}} &  \textit{train} &  6961 \\
        &   \textit{test} &   1741 \\
\hline
      total  &    &   8702 \\
\hline
\hline
        \multirow{2}{*}{\textbf{IR}} 
        &  \textit{corpus} &  309621 \\
        &   \textit{queries} &   10751 \\
\hline
\hline    
\end{tabularx}
    \caption{Summary of the PUGG dataset size.}
    \label{tab:statistics}
\end{table}

\section{Experimental Setup}
\label{sec:experimental-setup}

In this section, we outline the evaluation methodology used to assess the performance of baseline models on the PUGG dataset.

\paragraph{KBQA} For the KBQA baseline, we evaluated the performance of KAPING \cite{baek-etal-2023-knowledge-augmented}, a zero-shot framework that leverages an LLM for retrieving answer entities. We slightly modified the knowledge retriever module by incorporating a step that retrieves a subgraph of the KG by traversing $n$ edges, regardless of their direction, from the topic entities. Our preliminary experiments demonstrated enhanced performance of the modification, showcasing improvements in both accuracy and processing speed. Subsequently, we follow the original procedure, which involves retrieving $k$ triples based on their textual embeddings. For embedding purposes, we utilized the \textit{mmlw-retrieval-roberta-large} retrieval model \cite{dadas-etal-2024-pirb-comprehensive}. We employed \textit{gpt-3-turbo} as the LLM, prompted with tailored queries as detailed in Appendix \ref{sec:kbqa-prompts}. The hyperparameters were selected empirically, setting $k=40$ and choosing $n$ to be $3$ for Wikidata1H and $2$ for Wikidata2H. As a metric, we employed accuracy, which measures the proportion of answers included in the LLM's response for each question. It is calculated as follows: \[ \text{Accuracy} = \frac{1}{|Q|} \sum_{i=1}^{|Q|} \frac{\text{num of incl. answers}_i}{|\mathcal{A}_i|} \]

While \citet{baek-etal-2023-knowledge-augmented} also used accuracy, we refined it by calculating the correct answer proportion per example and excluding entities' aliases, providing a more realistic measure of KBQA efficacy.

\paragraph{MRC}

For the MRC task, we selected models commonly used for the extractive question answering task. We trained and evaluated HerBERT \cite{mroczkowski-etal-2021-herbert} models in an extractive fashion alongside a generative approach using the plT5 \cite{chrabrowa-etal-2022-evaluation} models. Models were trained for $10$ epochs and evaluated with SQuAD metrics \cite{rajpurkar-etal-2016-squad}. Exact match measures the percentage of predictions that exactly match the gold answer. The F1 metric measures the average token overlap between the prediction and ground truth answer, where both the prediction and answer are treated as a bag of tokens.

\paragraph{IR}
Recently, IR has gained significant interest within the Polish research community, and many models have been developed and are open to the research community. These models have already been pre-trained on large datasets, which is why we did not fine-tune them to our dataset. The silver retriever \cite{rybak2023silverretriever} model was trained on the MAUPQA dataset. 
We also evaluated E5 \cite{wang2024multilingual} multilingual embedding models, which were trained on contrastive objectives on large weakly-labeled text pairs and afterward fine-tuned on existing datasets and are performing very well on Polish texts. The MMLW retrieval models \cite{dadas-etal-2024-pirb-comprehensive} were trained on a parallel corpus with Polish-English text pairs with a \textit{bge-large-en} \cite{bge_embedding} teacher model and are currently on the top of the PIRB leaderboard. We also provide results of well-established BM25 \cite{bm25} baseline with Morfologik \footnote{\url{https://github.com/allegro/elasticsearch-analysis-morfologik}} plugin in Elasticsearch.

Additionally, we evaluated reranker models, focusing on those developed in the BEIR-PL benchmark and recent models that have appeared on the PIRB leaderboard. Those models were trained by \citealp{dadas2024assessing} with knowledge distillation from mT5-13B model introduced in the mMARCO publication \cite{bonifacio2021mmarco}. For reranking, we employed the BM25 retrieval algorithm to select the top $100$ passages for subsequent analysis. Finally, we provide a score of the combination of the best retriever and reranker, namely \textit{multilingual-e5-large retriever} and \textit{polish-reranker-large-ranknet reranker}, to evaluate currently the best IR pipeline available. We calculated the well-established metrics for the IR task: MRR@k, NDCG@k, Recall@k \cite{thakur2021beir,wojtasik2023beirpl}.

\section{Results and Discussion}
\label{sec:results-discussion}

\paragraph{KBQA}
The summarized results are presented in Table \ref{tab:kbqa-results}. For natural and template-based questions, utilizing KG significantly improves accuracy. The overall accuracy is not high, indicating the challenging nature of the newly introduced PUGG dataset. This complexity highlights its potential as a valuable resource for advancing research and development in the field of KBQA. As expected, reasoning over $1$-hop (1H) KG was easier than over $2$-hop (2H) KG, reflecting the increased complexity of KG. There is a clear gap in efficacy between natural and template-based questions. That was expected, as template-based questions were designed to be easier. Interestingly, they benefit more from the use of KG than the natural ones. We think that it can be caused by their schematic construction mechanism. Moreover, our pipeline for natural questions does not ensure the existence of appropriate reasoning paths in the graph, which could also cause lower efficacy.

\begin{table}[!htb]
    \centering
    \small
\begin{tabularx}{\linewidth}{XXXc}
\hline
         \textbf{Mode} &  \textbf{KG} &  \textbf{Retriever} &  \textbf{Accuracy} \\
\hline
\multicolumn{4}{c}{\textbf{KBQA (natural)}} \\
                w/o KG &        - &            - &     0.275 \\
                 w/ KG &        Wikidata1H &            3-hop &     0.342 \\
                 w/ KG &        Wikidata2H &            2-hop &     0.334 \\
    \hline
    \multicolumn{4}{c}{\textbf{KBQA (template-based)}} \\
         w/o KG &        - &            - &     0.210 \\
          w/ KG &        Wikidata1H &            3-hop &     0.674 \\
          w/ KG &        Wikidata2H &            2-hop &     0.669 \\
    \hline
    \multicolumn{4}{c}{\textbf{KBQA (all)}} \\
                    w/o KG &        - &            - &     0.250 \\
                     w/ KG &        Wikidata1H &            3-hop &     0.468 \\
                     w/ KG &        Wikidata2H &            2-hop &     0.461 \\
\hline
\end{tabularx}
    \caption{Results of the KBQA baselines.}
    \label{tab:kbqa-results}
\end{table}

\paragraph{MRC}

The results of the MRC baselines, as presented in Table \ref{mrc:results}, suggest that extractive models excel in identifying exact matches within the text. On the other hand, large generative models have demonstrated a capacity to achieve a high degree of general answer overlap, as reflected by their F1 scores. Compared to the baseline results disclosed in the PoQuAD publication \cite{poquad}, which reported exact match and F1 scores of $66.22$ and $81.39$, the current results suggest that the dataset constitutes a greater challenge for the models.

\begin{table}[!htb]
\centering
\small
\begin{tabularx}{\linewidth}{Xcc}
\hline
\textbf{Model name}         & \textbf{Exact Match} & \textbf{F1} \\ \hline
herbert-base-cased  & 42.91                & 66.41       \\
herbert-large-cased & \textbf{46.81}                & 70.42       \\ \hline
plt5-base           & 22.86                & 57.63       \\
plt5-large          & 38.88               &  \textbf{71.52}    \\
\hline
\end{tabularx}
\caption{Results of the MRC baselines. }
\label{mrc:results}
\end{table}

\paragraph{IR}

\begin{table*}[!htb]
\centering
\small
\begin{tabular}{ccccc}
\hline
\textbf{Model name }                                         & \textbf{NDCG@10 }             & \multicolumn{1}{l}{\textbf{MRR@10}} & \multicolumn{1}{l}{\textbf{Recall@10}} & \multicolumn{1}{l}{\textbf{Recall@100}} \\ \hline
\textbf{Retriever baselines}                        & \multicolumn{1}{l}{} & \multicolumn{1}{l}{}       & \multicolumn{1}{l}{}          & \multicolumn{1}{l}{}           \\ \hline
BM25                                                & 0.371                & 0.318                      & 0.549                         & 0.809                          \\
silver-retriever-base-v1.1                          & 0.523                & 0.457                      & 0.733                         & 0.923                          \\
mmlw-retrieval-roberta-base                         & 0.645                & 0.601                      & 0.805                         & 0.925                          \\
mmlw-retrieval-roberta-large                        & 0.700                & 0.653                      & 0.849                         & 0.946                          \\
multilingual-e5-base                                & 0.667                & 0.616                      & 0.828                         & 0.943                          \\
multilingual-e5-large                               & \textbf{0.741}                & \textbf{0.694}                      & \textbf{0.888}                         & \textbf{0.972}                          \\ \hline
\textbf{Retriever+Reranker baselines}               & \multicolumn{1}{l}{} & \multicolumn{1}{l}{}       & \multicolumn{1}{l}{}          & \multicolumn{1}{l}{}           \\ \hline
BM25+herbert-large-msmarco                          & 0.707                & 0.677                      & 0.797                         & 0.809                          \\
BM25+polish-reranker-base-ranknet                   & 0.701                & 0.671                      & 0.792                         & 0.809                          \\
BM25+polish-reranker-large-ranknet                  & 0.723                & 0.697                      & 0.802                         & 0.809                          \\
multilingual-e5-large+polish-reranker-large-ranknet & \textbf{0.813}                & \textbf{0.770}                      & \textbf{0.942}                         & \textbf{0.972}                          \\ \hline
\end{tabular}
\caption{Results of the IR baselines. The baselines are categorized into two groups: retriever baselines and retrievers with reranking baselines. For the reranking baselines, the top 100 retriever results undergo reranking.}
\label{ir:results}
\end{table*}

The scores presented in Table \ref{ir:results} reveal that the dataset poses a significant challenge for the lexical BM25 approach. The questions have limited lexical overlap; therefore, this method is ineffective. Nonetheless, current dense retrieval models are exhibiting high performance. Surprisingly, the \textit{mmlw-retrieval-roberta-large} model, despite being currently ranked at the top of the PIRB benchmark, still falls behind the \textit{multilingual-e5-large} model. This suggests that the dataset is a valuable resource for assessment and should be included in the PIRB benchmark in the future. The reranker models improved the BM25 rankings significantly, and combining a dense retriever with a reranker has achieved remarkably high scores across all metrics.

\section{Limitations and Future Work}
\label{sec:limitations-future-work}

This section outlines the limitations of our study and potential directions for future work. (1) The natural questions are open domain, focused on location and time, and are created and answered from the Polish cultural, political, and historical perspective. 
(2) The pipeline for natural questions may sometimes miss certain answer entities. This is because not all answers are present or explicitly referenced in the textual answer.
(3) Some of the KBQA natural questions might not have corresponding facts in the KG, as our pipeline does not guarantee the existence of an appropriate reasoning path between topic and answer entities. However, as Wikidata is continuously updated and expanded, this limitation may diminish in the future.
(4) The questions might contain grammatical imperfections or mental shortcuts yet remain understandable.
(5) Automated annotation with LLM led to variability in the precision of tagged answers in the MRC task due to the absence of specific tagging guidelines.
(6) Our study examined a limited number of baseline models. Future evaluations could, in particular, include open-source LLMs like Llama \citep{touvron2023llama} for MRC and KBQA tasks, as well as models that reason directly over the KG structure, such as PullNet \citep{sun-etal-2019-pullnet}, for KBQA task.
(7) While our focus was on standard tasks, we acknowledge the potential for exploring additional tasks using the PUGG dataset. These tasks include entity linking, subgraph retrieval, relation extraction, question type classification, and question generation.

\section{Conclusion}
\label{sec:conclusion}

To address the significant resource gap for low-resource languages, our work introduces the PUGG dataset, the first Polish KBQA dataset, which also encompasses MRC and IR tasks. It consists of natural and template-based factoid questions. The dataset is the outcome of our proposed semi-automated construction pipeline, designed for low-resource environments. Leveraging modern tools like LLMs as annotation assistants have significantly reduced the need for human labor. Additionally, we developed few utility methods, such as entity linking, which are useful in various contexts. The PUGG dataset and our pipeline's comprehensive implementation, findings, and detailed statistics from the PUGG dataset construction provide valuable insights for future research. Furthermore, the evaluation of baseline models on this dataset reveals its challenging nature, underscoring its potential to advance the field and contribute to developing more robust QA systems.

\section{Ethical considerations}

The process of dataset creation using LLMs and pre-existing datasets entails the potential risk of inheriting biases from both the models and the original data sources. To address this concern, a pipeline could incorporate multiple LLMs and diverse datasets as a mitigation strategy. 

We used sources with a low risk of containing private data or offensive content. However, during the human verification process, we further ensured that the dataset did not include such data. As mentioned in Section \ref{sec:pipeline-execution-human-ver-text-ans}, all annotators were employed in Poland and were fluent in Polish. They were familiar with the Polish culture and social context.

\section{Acknowledgments}
This work was supported by Polish Ministry of Education and Science under the programme: "Support for the participation of Polish scientific teams in international research infrastructure projects", agreement number 2024/WK/01 and project of the Minister of Digitization No. 1/WI/DBiI/202 (PLLUM). This work was also partially funded by the European Union under the Horizon Europe grant OMINO – Overcoming Multilevel INformation Overload (grant number 101086321, http://ominoproject.eu) co-financed with funds from the Polish Ministry of Education and Science under the programme entitled International Co-Financed Projects, grant no. 573977. 


\bibliography{custom}

\appendix

\section{Textual Answers Tagging}
\label{sec:appendix-tagging}

The designed prompt is presented in Table \ref{tab:kbqa-tagging}. The annotated spans were extracted from the LLM's responses using lemmatization and longest common sequence analysis.

\begin{table*}[!htb]
\centering
\begin{tabular}{p{0.02\textwidth} p{0.85\textwidth}}
\hline
\multicolumn{2}{c}{\textbf{Textual Answer Tagging Prompt}} \\
\hline
pl: &  \setbox0=\hbox{\begin{lstlisting}
 User:  
        Cytat to dokładna kopia tekstu słowo w słowo. Podam tobie tekst i pytanie. Twoim zadaniem będzie znalezienie w tekście DOKŁADNEGO cytatu. Cytat musi być najbliższy odpowiedzi lub taki, który może być potencjalną odpowiedzią. Musi to być najkrótszy możliwy cytat w tekście. Nie należy zmieniać żadnych słów. Nie odmieniaj słów. Nie dodawaj żadnych dodatkowych słów, abym mógł go skopiować. Więc proszę nie zmieniać nawet kapitalizacji.
 Assistant:
        Jasne, przytoczę tylko dokładny cytat. Nie będę dodawał żadnych słów. Nie będę zmieniał słów. Nie będę zmieniał przypadków słów. Nie zmienię wielkości liter.
 User:
        Context: "[START]Elżbieta II (; ur. 21 kwietnia 1926 w Londynie, zm. 8 września 2022 w Balmoral) - królowa Zjednoczonego Królestwa Wielkiej Brytanii i Irlandii Północnej z dynastii Windsorów od 6 lutego 1952 (koronowana 2 czerwca 1953) do 8 września 2022.[END]" 
        Question: w którym roku urodziła się królowa elżbieta ii?
        A: "
 Assistant:
        21 kwietnia 1926"
 User: 
        Context: "[START]{context}[END]" 
        Question: {question}
        A: "
\end{lstlisting}}\mbox{}\box0 \\
\hline
en: &  \setbox0=\hbox{\begin{lstlisting}
 User:  
        A quote is an exact copy of the text word for word. I will give you the text and the question. Your task will be to find the EXACT quote in the text. The quote must be the closest to the answer or one that could be a potential answer. It must be the shortest possible quote in the text. Do not change any words. Do not inflect words. Do not add any additional words so that I can copy it. So please don't even change the capitalization.
 Assistant:
        Sure, I will just quote the exact quote. I will not add any words. I will not change the words. I will not change the word cases. I will not change the case of the letters.
 User:
        Context: "[START]Elizabeth II (; born April 21, 1926 in London, died September 8, 2022 in Balmoral) - Queen of the United Kingdom of Great Britain and Northern Ireland of the Windsor dynasty from February 6, 1952 (crowned June 2, 1953) to September 8, 2022.[END]" 
        Question: in what year was Queen Elizabeth ii born?
        A: "
 Assistant:
        April 21, 1926"
 User: 
        Context: "[START]{context}[END]" 
        Question: {question}
        A: "
\end{lstlisting}}\mbox{}\box0 \\
\hline
\end{tabular}
\caption{Textual answer tagging prompt. The prompt was translated to English for non-Polish readers; it was not used or tested in this form.}
\label{tab:kbqa-tagging}
\end{table*}

\section{Topic Entity Linking}\label{sec:appendix-entity-linking}

The designed entity linking method primarily relies on the Wikipedia search engine, title similarity, and information about the \textit{neighborhood of the question}.

The \textbf{Wikipedia search engine} is accessed via the MediaWiki API \footnote{\url{https://www.mediawiki.org/wiki/API:Search}}. This search system identifies page titles or content that match a given textual query. \textbf{Title similarity} is measured by assessing the similarity of provided texts, utilizing both the longest common sequence and the longest common prefix approaches. To construct the \textbf{neighborhood of the question}, we retrieved Wikipedia pages from the top $10$ Google search results and then extracted the first five links from each of these articles. These results are then used to determine whether the entity found by the algorithm belongs to such a neighborhood. It is important to note that, in this context, \textit{the neighborhood of the question} is not associated with the KG.

As the output, we expect four types of entities: exact entities, neighborhood entities, named entities, and combined entities. Detailed information on this process can be found in the pseudocode provided in Algorithm \ref{alg:algorithm1}. For tokenization, lemmatization, and NER, we used the SpaCy tool \cite{honnibal2020spacy} with the \textit{pl\_core\_news\_lg} model.

\begin{algorithm}
\caption{Entity Linking Method}
\label{alg:algorithm1}
\textbf{Input:} 

\quad $Q$ - input question.

\textbf{Constants:}

\quad $ L \gets [$noun, adjective, proper noun, unknown]

\quad $T \gets$ tokenize\_to\_words($Q$)

\quad $N \gets$ named\_entities($Q$)

\textbf{Output:}

\quad ${E\textsubscript{exact}}$ - set of entities closely matching the title of Wikipedia pages

\quad ${E\textsubscript{nbhd}}$ - set of entities not precisely matching Wikipedia titles but belonging to the question neighborhood

\quad ${E\textsubscript{named}}$ - set of named entities belonging to the question neighborhood

\quad ${E\textsubscript{comb}}$ - set of entities formed by combining two or more words \\

\textbf{Algorithm:}
\begin{algorithmic}[htb]
\For{each $t \in T$}
    \If{$\text{pos}(t) \in L$}
        \State $res \gets \text{search\_wikipedia}(t)$
        \State $l \gets \text{lemma}(t)$
        \State $E_{exact} \gets \text{high\_similarity}(res, l)$
    \EndIf
\EndFor

\For{each $n \in N$}
    \State $res \gets \text{search\_wikipedia}(n)$
    \State $E_{named} \gets \text{in\_neighborhood}(res)$
\EndFor

\For{each $t \in T$}
    \If{$\text{pos}(t)$ in $L$}
        \State $res \gets \text{search\_wikipedia}(n)$
        \State $E_{nbhd} \gets \text{in\_neighborhood}(res)$
    \EndIf
\EndFor

\For{each $t \in T$}
    \If{$\text{pos}(t) == \text{'noun'}$}
        \State $R \gets \text{get\_nouns}(\text{children}(t))$
        \State $A \gets \text{get\_adjectives}(\text{children}(t))$

        \State $R_q \gets R \times [t]$

        \State $A_q \gets A \times [t]$

        \For{each $q \in R_q \cup A_q$}
            \State $res \gets \text{search\_wikipedia}(q)$
            \State $E_{comb} \gets \text{in\_neighborhood}(res)$
        \EndFor
    \EndIf
\EndFor

\end{algorithmic}
\end{algorithm}

\section{Human Verification} \label{sec:appendix-human-verification}

\subsection{First Stage}

To ensure high-quality data, the annotation team included annotators and a super-annotator. The process involved: (1) initial guideline preparation, (2) a full review of annotator decisions reviewed by the super-annotator, and (3) a targeted review of problematic examples by the super-annotator. This process refined the guidelines and focused on resolving ambiguities in annotations.
Examples with improperly formulated questions or lacking information for accurate answers were rejected, especially those with significant grammatical or lexical errors that made them incomprehensible.
Technically, this step involved flagging documents in the Inforex system \cite{marcinczuk-etal-2017-inforex}, with the following set of flags:
(1) \textit{correct}: indicates both the question and answer are correct in the passage.
(2) \textit{incorrect question}: indicates the question is formulated incorrectly.
(3) \textit{incorrect passage}: indicates the passage does not answer the question.
(4) \textit{incorrect fragment}: indicates the answer is located elsewhere in the passage.

\subsection{Second Stage}

Two annotators carried out this stage. To facilitate a consistent and measurable approach, we separated $10\%$ of the examples as common for both annotators, while the rest were individually assigned. These shared examples served as a basis for calculating annotation metrics and ensuring reliability and consistency in the annotation process.
Annotating the correct answers was a straightforward task. However, the annotation of topic entities presented more complexity. As \citet{Rosales-MendezP18} have pointed out, there is no consensus on the concept of an entity and what entity linking should link to, as it varies greatly depending on the application. Due to the absence of universally acknowledged guidelines, we defined a topic entity as a source entity from which the reasoning method should begin its process. In cases where annotators were uncertain about either answer or topic entities, the problematic examples were rejected to maintain the dataset's quality. The entire second stage of the annotation process was carried out using a spreadsheet application. During the annotation of answer entities and topic entities, we achieved Cohen's kappa scores of $0.785$ and $0.675$, with accuracy scores of $0.892$ and $0.895$, respectively.

\section{Template-based KBQA}
\subsection{Templates}
\label{sec:appendix-templates}

We have developed $8$ templates for schematic question creation, detailed in Table \ref{tab:templates}. We distinguish the following three general techniques.

\textbf{N-hop templates} retrieve information by traversing $N$ relations from the given entity.

\textbf{Reverse N-hop templates} function similarly but involve traversing in the reverse direction.

The \textbf{Entity Mask} technique enriches questions by referring to the answer without direct mention. For example, instead of naming "Ludwig van Beethoven", we might use "composer".

\begin{table*}[!htb]
\centering
\tiny
\begin{tabular}{|l|l|l|l|l|}
\hline
\multirow{ 2}{*}{\textbf{Template name}} &  & \textbf{Natural Language} & \multirow{ 2}{*}{\textbf{Examples}} & \multirow{ 2}{*}{\textbf{SPARQL Template}}\\
&  & \textbf{Template} & &\\
\hline
One-hop & pl & Jakie ... ma ...? &  Q: Jakie \{imię\} ma \{Ludwig van Beethoven\}? & SELECT ?answerEntity \\
&  &  &   A: \{Ludwig\}. & WHERE \{\{ \\
\cline{2-4}
& en & What is the ... of ...? & Q: What is the \{given name\} of \{Ludwig van &  \quad  wd:Q255 wdt:P735 ?answerEntity. \\
&  & &  Beethoven\}? & \}\}  \\
&  & &   A: \{Ludwig\}. & \\
\hline
One-hop with & pl & Jak nazywał się ..., &  Q: Jak nazywał się \{metropolia\}, które jest \{miejsce   & SELECT ?answerEntity \\
entity mask & & które jest ... ...? & śmierci\} \{Ludwig van Beethoven\}?  & WHERE \{\{ \\
& & & A: \{Wiedeń\}. & \quad wd:Q255 wdt:P20  ?answerEntity.\\
\cline{2-4}
& en & What was the name & Q: What was the name of the \{metropolis\}, which is the & \quad ?answerEntity wdt:P31 wd:Q200250. \\
& &  of the ..., which is & \{place of death\} of \{Ludwig van Beethoven\}?  & \}\} \\
& &  the ... of ...? & A: \{Vienna\}. & \\
\hline
Two-hop & pl & Jakie ... ma ... ...? & Q: Jakie \{obywatelstwo\} ma \{matka\} \{Ludwig & SELECT ?answerEntity \\
& & & van Beethoven\}? & WHERE \{\{\\
& & & A: \{Niemcy\}. & \quad wd:Q255 wdt:P25 ?relatedEntity.\\
\cline{2-4}
& en & What is the ... of ...'s  & Q: What is the \{country of citizenship\} of & \quad ?relatedEntity wdt:P27 ?answerEntity. \\
& & ...? & \{Ludwig van Beethoven\}'s \{mother\}? & \}\} \\
& & & A: \{Germany\}. & \\
\hline
Reverse one-hop  & pl & Czyim ... jest ...?& Q: Czyim \{student\} jest \{Carl Czerny\}? & SELECT ?answerEntity  \\ 
& & & A: \{Ludwig van Beethoven, Antonio Salieri\}. & WHERE \{\{ \\
\cline{2-4}
& en & Whose ... is ...? & Q: Whose \{student\} is \{Carl Czerny\}? & \quad ?answerEntity wdt:P802 wd:Q215333. \\
& & & A: \{Ludwig van Beethoven, Antonio Salieri\}. & \}\} \\
\hline
Reverse one-hop  & pl & Jak nazywał się ..., & Q: Jak nazywał się \{kompozytor\}, którego  & SELECT ?answerEntity \\ 
with mask entity &  & którego ... jest ...? & \{rodzeństwo\} jest \{Kaspar Anton Karl van & WHERE \{\{ \\
& & & Beethoven\}? & \quad ?answerEntity wdt:P3373 wd:Q6374627.\\
& & & A: \{Ludwig van Beethoven\}. & \quad ?answerEntity wdt:P106 wd:Q36834.  \\
\cline{2-4}
& en & What was the name & Q: What was the name of the \{composer\} whose &  \}\} \\
& & of the ...  whose ... & \{sibling\} is \{Kaspar Anton Karl van Beethoven\}? & \\
& & is ...? & A: \{Ludwig van Beethoven\}. & \\
\hline
Reverse two-hop  & pl & Czyim ... jest ..., & Q: Czyim \{student\} jest \{Ferdinand Ries\}, a & SELECT ?answerEntity \\ 
& & a ... jest ...? & \{nauczyciel\} jest \{Joseph Haydn\}? & WHERE \{\{ \\
& & & A: \{Ludwig van Beethoven\}. & \quad ?answerEntity wdt:P802 wd:Q213558.\\
\cline{2-4}
& en & Whose ... is ..., and & Q: Whose \{student\} is \{Ferdinand Ries\}, and & \quad ?answerEntity wdt:P1066 wd:Q7349.\\
& & ... is ...?  & \{teacher\} is \{Joseph Haydn\}? & \}\}\\
& & & A: \{Ludwig van Beethoven\}. & \\
\hline
Reverse two-hop  & pl & Jak nazywał się ..., & Q: Jak nazywał się \{kompozytor\}, którego & SELECT ?answerEntity  \\
with mask entity & & którego ... jest ..., a & \{przyczyna śmierci\} jest \{marskość wątroby\}, & WHERE \{\{ \\
& & którego ... jest ...? & a którego \{miejsce śmierci\} jest \{Wiedeń\}? & \quad ?answerEntity wdt:P509 wd:Q147778.\\
& & & A: \{Ludwig van Beethoven\}. & \quad ?answerEntity wdt:P20 wd:Q1741.\\
\cline{2-4}
& en & What was the name & Q: What was the name of the \{composer\} & \quad ?answerEntity wdt:P106 wd:Q36834.\\
& & of the ... whose ... & whose \{cause of death\} is \{cirrhosis of the liver\}, & \}\} \\
& & is ... and whose ... & and whose \{place of death\} is \{Vienna\}? & \\
& &  is ...? & A: \{Ludwig van Beethoven\}. &\\
\hline
Mixed & pl & Jakie ... ma ..., & Q: Jakie \{miejsce urodzenia\} ma \{kompozytor\}, & SELECT ?answerEntity \\
& & którego ... jest ...? & którego \{ojcem\} jest \{Johann van Beethoven\}? & WHERE \{\{ \\
& & & A: \{Bonn\}. & \quad?relatedEntity wdt:P106 wd:Q36834.\\
\cline{2-4}
& en & What is the ... of & Q: What is the \{place of birth\} of the \{composer\} & \quad ?relatedEntity wdt:P22 wd:Q2153541.\\
& & the ... whose ... & whose \{father\} is \{Johann van Beethoven\}? & \quad ?relatedEntity wdt:P19 ?answerEntity.\\
& & is ...? & A: \{Bonn\}. & \}\} \\
\hline
\end{tabular}
\caption{The question templates used for template-based questions. The English example data is presented for non-Polish readers, but the pipeline was originally executed on Polish data for PUGG creation.}
\label{tab:templates}
\end{table*}

\subsection{Paraphrasing and Inflection Prompts}
Table \ref{tab:inflection-paraphrasing} presents the prompts for inflecting and paraphrasing questions constructed using the natural language templates.


\begin{table*}[!htb]
\centering
\begin{tabular}{p{0.02\textwidth} p{0.85\textwidth}}
\hline
\multicolumn{2}{c}{\textbf{Inflection Prompt}} \\
\hline
pl: &  \setbox0=\hbox{\begin{lstlisting}
 User:  
        Zmień błędne końcówki wyrazów w pytaniu.  Pamiętaj, że nie wolno zmieniać podstaw słów, zastępować ich synonimami ani dodawać nowych. Nie można zmieniać kolejności słów.
 Assistant:
        Jasne, poprawię błędne końcówki wyrazów w pytaniu. Nie będę zmieniał kolejności słów. Nie będę dodawał nowych słów. Nie będę zastępował synonimami.
 User:
        "Czyim dzieci jest Maria Gorecka?"
 Assistant:
        "Czyim dzieckiem jest Maria Gorecka?"
 User:
        "Jak nazywał się gmina miejska w Niemczech, który jest miejsce pobytu Adam Mickiewicz?"
 Assistant:
        "Jak nazywała się gmina miejska w Niemczech, która była miejscem pobytu Adama Mickiewicza?"
 User:
        "{question}"
\end{lstlisting}}\mbox{}\box0 \\
\hline
en: &  \setbox0=\hbox{\begin{lstlisting}
 User:
        Change the incorrect word endings in the question. Remember not to change the base words, replace them with synonyms, or add new ones. You cannot change the word order.
 Assistant:
        Sure, I will correct the incorrect word endings in the question. I will not change the word order. I will not add new words. I will not replace them with synonyms.
 User:
        "Whose children is Maria Gorecka?"
 Assistant: 
        "Whose child is Maria Gorecka?"
 User:
        "What was the name of the urban municipality in Germany, which is the residence of Adam Mickiewicz?"
 Assistant: 
        "What was the name of the urban municipality in Germany, which was the residence of Adam Mickiewicz?"
 User:
        "{question}"
\end{lstlisting}}\mbox{}\box0 \\
\hline
\multicolumn{2}{c}{\textbf{Paraphrasing Prompt}} \\
\hline
pl: &  \setbox0=\hbox{\begin{lstlisting}
 User:  
        Proszę, przeformułuj następujące pytanie, zachowując jego sens.
 Assistant:
        Jasne, zrobię to, nie zmieniając sensu pytania.
 User:
        "Czyim dzieckiem jest Maria Gorecka?"
 Assistant:
        "Kim są rodzice Marii Goreckiej?"
 User:
        "{question}"
\end{lstlisting}}\mbox{}\box0 \\
\hline
en: &  \setbox0=\hbox{\begin{lstlisting}
 User:  
        Please, paraphrase the following question while maintaining its meaning.
 Assistant:
        Sure, I'll do that without changing the question's meaning.
 User:
        "Whose child is Maria Gorecka?"
 Assistant: 
        "Who are the parents of Maria Gorecka?"
 User:
        "{question}"
\end{lstlisting}}\mbox{}\box0 \\
\hline
\end{tabular}
\caption{Inflection and paraphrasing prompts used for template-based KBQA. The prompts were translated to English for non-Polish readers; they were not used or tested in this form.}
\label{tab:inflection-paraphrasing}
\end{table*}

\subsection{Human Verification}

Inflected and paraphrased questions were verified using the following set of annotation flags: \textit{correct}, \textit{incorrect}, and \textit{resembling}.

\textbf{Correct} implies the semantic meaning of the processed question remains unchanged compared to the original. \textbf{Incorrect} flags a change in semantic meaning. For instance, the original question \textit{'Who is the creator of the web browser?'} paraphrased as \textit{'What material is the web browser created of?'} illustrates this change. It is also worth mentioning that incorrect questions often involve the reversal of relations: \textit{Whose doctoral supervisor is Max Perutz?} was paraphrased as \textit{Who is Max Perutz's doctoral supervisor?}. The fact that LLMs may struggle to understand reverse connections, was also highlighted by \citet{berglund2023reversal}. During annotation, we noticed some question patterns frequently repeated in specific templates like one-hop templates. We labeled these as \textbf{resembling} and excluded them from the final dataset. For example, \textit{'Where was X born?'}, was common due to the \textit{'place of birth'} being a prevalent relation for people on Wikidata. The statistics of verification are presented in Table \ref{tab:annotation_statistics}.

\begin{table}[!htb]
\centering
\begin{tabular}{lrrr}
\hline
\textbf{Template name} & \textbf{C}& \textbf{I} & \textbf{R} \\ 

\hline
One Hop                       &   137    &  393          & 89   \\ 
One Hop With Mask           &   185    & 335          & 69   \\ 
Two Hop                        &   301    &  290          & 0   \\ 
Reverse One Hop               &   307    &  176           & 0   \\ 
Reverse One Hop W/ Mask    &   220    &  312           & 0   \\
Reverse Two Hop               &   398    &  88           & 0     \\ 
Reverse Two Hop W/ Mask   &   167    &  275          & 34   \\ 
Mixed                         &   231    &  224           & 0   \\ 
\hline
\end{tabular}

\caption{The number of \textbf{c}orrect, \textbf{i}ncorrect, and \textbf{r}esembling questions according to the manual verification for template-based questions.}
\label{tab:annotation_statistics}
\end{table}

\section{Detailed statistics} \label{sec:appendix-detailed-stats}

\subsection{Pipeline stages}

Table \ref{tab:detailed-statistics} summarizes the number of examples processed at each stage of the PUGG construction, both for natural and template-based questions. Each step either increased or reduced the number of examples. This step-by-step analysis could be beneficial for scientists and engineers aiming to execute similar pipelines. It offers a precise estimate of the volume of data necessary at the beginning and the anticipated human labor required during the verification stages. Notably, textual answer tagging and entity verification stages contribute to the most significant reductions in data volume. 

The initial steps (gathering questions from existing QA datasets, extracting prefixes, and formulating questions) significantly increased the number of potential examples. This increase was essential for the subsequent stages that reduced questions. The detailed reasons for the reductions are described in Section \ref{sec:pipeline} and \ref{sec:pipeline-execution}, however they are summarized in the following points.

\paragraph{Natural Questions}
(1) Questions for textual answer tagging: questions without any Wikipedia article in the top 10 results from the search engine were discarded.
(2) Questions for textual answer tagging: reduced to those where tags generated by the LLM were successfully parsed.
(3) Questions with the correct passage: filtered to questions with passages correctly answered the questions.
(4) Correct textual answers: filtered to questions with correct textual answers.
(5) Questions with verified answer entities: questions without any correct answer entities were discarded.
(6) Questions with verified topic entity: questions without any correct topic entities were discarded.
(7) KBQA/MRC examples: the final dataset examples differ from those in the corresponding previous steps due to several manual interventions. These include deduplication and manual entity linking.

\paragraph{Template-based Questions}
(1) After filtering: questions without high similarity to their original form were filtered out.
(2) After verification: human verification ensured the meaningfulness of questions.

\begin{table}[!htb]
    \centering
\begin{tabular}{lS[table-format=5.0]}
\hline
                                      \textbf{Data} &   \textbf{\#} \\
\hline
\multicolumn{2}{c}{\textbf{Natural}} \\
       Questions from existing QA datasets & 17019 \\
                                  Extracted Prefixes & 33467 \\
                      Formulated questions & 90666 \\
                        Retrieved Wikipedia articles & 18055 \\
              Questions for textual answer tagging & 31780 \\
                  Questions with successfully parsed tag & 19296 \\
                          Questions with correct passage &  10751 \\
                   Questions with correct textual answer &  8772 \\
Questions with verified answer entities &  3832 \\
 Questions with verified topic entities  &  3509 \\
   KBQA examples &  3471 \\
    MRC examples &  8702 \\
     IR examples &  10751 \\
\hline
\multicolumn{2}{c}{\textbf{Template-based}} \\
Executed templates & 14400 \\
After filtering & 4231 \\
After verification & 2122 \\

\hline
\end{tabular}
    \caption{Detailed statistics of the executed pipelines: natural and template-based.}
    \label{tab:detailed-statistics}
\end{table}

\subsection{Outcome}

Table \ref{tab:detailed-statistics-outcome} provides detailed statistics of PUGG, including unique topics, answers, and relations for both natural and template-based questions. Table \ref{tab:detailed-statistics-outcome-tamplates} shows the distribution of examples across different template types used in the template-based questions.

\begin{table}[!htb]
    \centering
    \begin{tabular}{lS[table-format=4.0]S[table-format=4.0]S[table-format=4.0]}
    \hline
        \multirow{ 2}{*}{\textbf{Subset}} & \textbf{\# unique} & \textbf{\# unique} & \textbf{\# unique} \\ 
        & \textbf{topics} & \textbf{answers} & \textbf{relations} \\ 
        \hline
        \multicolumn{4}{c}{\textbf{Natural}} \\
        train & 1985 & 3563 & – \\ 
        test & 610 & 1148 & – \\ 
        \hline
        \multicolumn{4}{c}{\textbf{Template-based}} \\
        train & 1787 & 1783 & 125 \\ 
        test & 537 & 573 & 91 \\ 
        \hline
    \end{tabular}
    \caption{Summary of unique topics, answers, and relations in the training and test sets for both natural and template-based questions. Note that we do not provide the number of relations in the natural dataset because, due to the construction pipeline characteristics, we do not know the exact reasoning path.}
\label{tab:detailed-statistics-outcome}
\end{table}

\begin{table}[!htb]
    \centering
    \begin{tabular}{lrr}
    \hline
        \textbf{Template name} & \textbf{Train} & \textbf{Test} \\ \hline
        One Hop & 311 & 82 \\ 
        One Hop With Mask & 261 & 74 \\ 
        Two Hop & 229 & 60 \\ 
        Reverse One Hop & 134 & 50 \\ 
        Reverse One Hop With Mask & 279 & 55 \\ 
        Reverse Two Hop & 68 & 20 \\ 
        Reverse Two Hop With Mask & 230 & 45 \\ 
        Mixed & 185 & 39 \\ \hline
    \end{tabular}
    \caption{Distribution of train and test examples across different template types in the constructed template-based question set.}
\label{tab:detailed-statistics-outcome-tamplates}
\end{table}

\section{KBQA Baseline Prompts} \label{sec:kbqa-prompts}

We adapted the LLM prompt from KAPING \cite{baek-etal-2023-knowledge-augmented} by translating and slightly modifying it to emphasize the need for listing entities in their non-inflected form. The adapted prompt is presented in Table \ref{tab:kbqa-prompt}. 

\begin{table*}[!htb]
\centering
\begin{tabular}{p{0.02\textwidth} p{0.85\textwidth}}
\hline
\multicolumn{2}{c}{\textbf{KBQA Baseline Prompt (w/o KG)}} \\
\hline
pl: &  \setbox0=\hbox{\begin{lstlisting}
Pytanie: {question}
Encje które są odpowiedzią: 
\end{lstlisting}}\mbox{}\box0 \\
\hline
en: &  \setbox0=\hbox{\begin{lstlisting}
Question: {question}
Entities which are the answer: 
\end{lstlisting}}\mbox{}\box0 \\
\hline
\multicolumn{2}{c}{\textbf{KBQA Baseline Prompt (w/ KG)}} \\
\hline
pl: &  \setbox0=\hbox{\begin{lstlisting}
Poniżej znajdują się fakty w postaci trójek grafu wiedzy w formacie (encja, relacja, encja), mające znaczenie do udzielenia odpowiedzi na pytanie.
{triples}
Pytanie: {question}
Encje które są odpowiedzią: 
\end{lstlisting}}\mbox{}\box0 \\
\hline
en: &  \setbox0=\hbox{\begin{lstlisting}
Below are facts in the form of knowledge graph triples in the format (entity, relation, entity), relevant to answering the question.
{triples}
Question: {question}
Entities which are the answer: 
\end{lstlisting}}\mbox{}\box0 \\
\hline

\end{tabular}
\caption{KBQA baseline Prompts. The prompts were translated to English for non-Polish readers; they were not used or tested in this form.}
\label{tab:kbqa-prompt}
\end{table*}

\end{document}